\documentclass[letterpaper, 10 pt, conference]{ieeeconf}  % Comment this line out if you need a4paper
\usepackage{xcolor}
\usepackage{amsmath}

 \usepackage[separate-uncertainty=true, multi-part-units=single, product-units=single]{siunitx}
\usepackage{graphicx}
\usepackage{subcaption}
\usepackage{hyperref}

\IEEEoverridecommandlockouts                              % This command is only needed if 
\title{\LARGE \bf
Towards Motion Compensation in Autonomous Robotic Subretinal Injections}

\author{Demir Arikan$^{1,2}$,  Mojtaba Esfandiari$^{2}$, Peiyao Zhang$^{2}$, Michael Sommersperger$^{1}$, Shervin Dehghani$^{1}$,\\Russel H. Taylor$^{2}$,  M. Ali Nasseri$^{1,4}$, Peter Gehlbach$^{3}$, Nassir Navab$^{5}$ and Iulian Iordachita$^{2}$%
\thanks{
*This work is supported by the U.S. National Institutes of Health under the grant numbers R01EB023943, R01EB025883, R01EB34397, and partially by JHU internal funds.}%
\thanks{$^{1}$D. Arikan, M. Sommersperger, S. Dehghani, M. Ali Nasseri are with Department of Computer Science, Technische Universit{\"a}t M{\"u}nchen, Munich 85748 Germany {\tt\small demir.arikan@tum.de}}%
\thanks{$^{2}$D. Arikan, M. Esfandiari, P. Zhang, R. H. Taylor and I. Iordachita are with the Laboratory for Computational Sensing and Robotics, Johns Hopkins University, Baltimore, MD, USA}%
\thanks{$^{3}$P. Gehlbach is with Wilmer Eye Institute, Johns Hopkins Hospital, Baltimore, MD, USA}%
\thanks{$^{4}$M. Ali Nasseri is with Augenklinik und Poliklinik, Klinikum rechts der Isar der Technische Universit{\"a}t M{\"u}nchen, M{\"u}nchen 81675 Germany}
\thanks{$^{5}$N. Navab is a full professor and head of the Chair for Computer Aided Medical Procedures \& Augmented Reality, Technical University of Munich, 85748 Munich, Germany}
}

\begin{document}

\maketitle
\thispagestyle{empty}
\pagestyle{empty}

%%%%%%%%%%%%%%%%%%%%%%%%%%%%%%%%%%%%%%%%%%%%%%%%%%%%%%%%%%%%%%%%%%%%%%%%%%%%%%%%
\begin{abstract}
Exudative (wet) age-related macular degeneration (AMD) is a leading cause of vision loss in older adults, typically treated with intravitreal injections. 
Emerging therapies, such as subretinal injections of stem cells, gene therapy, small molecules and RPE cells require precise delivery to avoid damaging delicate retinal structures. 
Robotic systems can potentially offer the necessary precision for these procedures. 
This paper presents a novel approach for motion compensation in robotic subretinal injections, utilizing real time Optical Coherence Tomography (OCT). 
The proposed method leverages B$^{5}$-scans, a rapid acquisition of small-volume OCT data, for dynamic tracking of retinal motion along the Z-axis, compensating for physiological movements such as breathing and heartbeat. 
Validation experiments on \textit{ex vivo} porcine eyes revealed challenges in maintaining a consistent tool-to-retina distance, with deviations of up to 200 \textmu m for 100 \textmu m amplitude motions and over 80 \textmu m for 25 \textmu m amplitude motions over one minute. 
Subretinal injections faced additional difficulties, with phase shifts causing the needle to move off-target and inject into the vitreous.
These results highlight the need for improved motion prediction and horizontal stability to enhance the accuracy and safety of robotic subretinal procedures.

\end{abstract}

%%%%%%%%%%%%%%%%%%%%%%%%%%%%%%%%%%%%%%%%%%%%%%%%%%%%%%%%%%%%%%%%%%%%%%%%%%%%%%%%
\section{INTRODUCTION}

\begin{figure*}
    \centering
    \includegraphics[width=\linewidth]{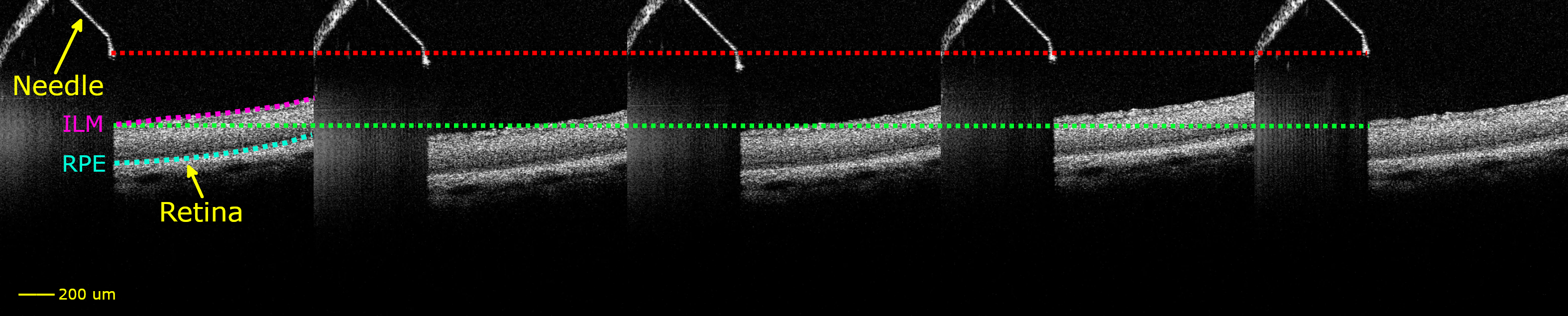}
    \caption{OCT B-scan images showing the simulated breathing motion transmitted to the retina with the needle following the same motion while trying to keep a constant distance from it. 
    Red and green lines represent the starting positions of the needle tip and retina respectively. 
    Our motion compensation is based on comparing the current with the previously acquired volumes, so we see a latency in the needle's motion (scans 3 and 5). The needle continues to move in the direction opposite the retina. 
    The pink and cyan lines are the internal limiting membrane (ILM) and retinal pigment epithelium (RPE) layers of the retina respectively.}
    \label{fig:breathing-phases}
\end{figure*}

About 8.7$\%$ of global blindness is caused by age-related macular degeneration (AMD), the leading cause of vision loss in individuals over 60 \cite{wong2008natural,kawasaki2010prevalence,mitchell1995prevalence,klein1999prevalence,klaver2001incidence}. With the rapid aging of the population, the prevalence of AMD is expected to rise from 196 million in 2020 to 288 million by 2040 \cite{wong2014global}. 
AMD is caused by multiple factors that, including senescence, leading to retinal pigment epithelial (RPE) cell degeneration and, eventually, loss of photoreceptors. 

Currently, the primary treatment for wet AMD involves injecting anti-VEGF (vascular endothelial growth factor) drugs directly into the vitreous cavity \cite{finger2020anti}. This symptomatic treatment delays disease progression but frequent intravitreal injections are costly and inconvenient to the patients and have serious risks, such as endophthalmitis \cite{day2011ocular}. 

Recent techniques such as transplantation of stem cells and injection of gene therapy vectors are solutions to treat AMD effectively. These approaches treat the cause of the issue through direct delivery of therapeutic agents into the subretinal space \cite{rakoczy2017gene, zhao2017development}. 
Each of these methods requires subretinal injections. 
Since photoreceptor and RPE cells are delicate and non-regenerative, any damage can result in irreversible vision loss. 
Therefore, a safe and efficient subretinal injection necessitates a precise and adaptable-depth insertion of a microsurgical cannula through the retina’s internal limiting membrane (ILM) and careful positioning of its tip between the photoreceptor and RPE cells \cite{ochakovski2017retinal, irigoyen2022subretinal}. 

Freehand subretinal injection remains challenging due to human physiological limits, including hand tremor, which is about 182 \textmu m \cite{riviere2000study}, making precise needle tip positioning extremely difficult inside the, on average 250 \textmu m thick, retina. Another major challenge is the lack of proper needle tip visualization relative to the target area \cite{zhao2017development}. 

Several surgical robotic platforms have been developed, including the Steady Hand Eye Robot (SHER) \cite{uneri2010new}, Intraocular Robotic Interventional and Surgical System (IRISS) \cite{rahimy2013robot}, Preceyes Surgical System \cite{van2009design}, to name a few \cite{gijbels2014experimental, jingjing2014design, nasseri2013introduction}. All these platforms filter out physiological hand tremor.
The SHER is a 5-degree-of-freedom (DoF) surgical robotic system developed at Johns Hopkins University. 
Several versions of these robots have been built, such as SHER 2.0 and SHER 2.1, which are serial manipulators and SHER 3.0, which has a compact parallel mechanism \cite{zhao2023human}. 
These robots are typically controlled by a cooperative strategy with an admittance-based control algorithm in which the force applied by the user's hand to the robot handle is measured by a force/torque sensor and interpreted as the desired end-effector velocity of the robot \cite{esfandiari2024cooperative, Henry2025-dg}. 
Cooperative control strategy involves direct visualization of the surgical site through a microscope followed by manual guidance of the robotic tool, which is a challenging task due to constrained visualization in a submillimeter workspace. 
Prior work has been performed to autonomously navigate the robotic system to the target area using deep learning methods integrated with intraoperative optical coherence tomography (iOCT) imaging, which provides high-resolution cross-sectional 2D B-scans or volumetric 3D C-scans of the retinal layers, enabling submillimeter precision for robotic tool placement and control \cite{ zhou2019towards, sommersperger2021real, mach2022oct, dehghani2023robotic}. 

Although these methods demonstrate promising outcomes in guiding the cannula along a pre-planned path to a target site in \textit{ex vivo} porcine eye experiments, they do not account for dynamic tissue deformation during insertion. Tool-tissue interaction during subretinal injection causes deformation in the retina, dynamically shifting the target point. 
Hence, a pre-planned trajectory with a fixed target point is not a realistic assumption and cannot precisely guide the cannula tip to a desired target. Of note, shifts in the target point along the $X$ and $Y$ axes are less concerning, but precise control over the $Z$-axis (perpendicular to the retina layer) remains essential. 

Our prior work, addressed the retinal tissue deformation along the $Z$-axis by real time feedback of the tissue deformation and the target point movement. We used our novel OCT-based tracking method and studied its feasibility in an \textit{ex vivo} porcine eye experiment \cite{arikan2024}. 
This depth accuracy ensures that injections consistently reach the correct tissue layer, especially near the RPE, where maintaining a specific depth is critical to delivering effective treatment without risking tissue damage. 
This approach represents an initial advancement in robotic subretinal injection adaptable to retina deformation; by creating and regularly updating a virtual target layer that is defined at a relative depth between the ILM and RPE layers, it enhances procedural safety and ensures that the fragile RPE cells remain intact during the insertion. 
We placed the porcine eyes on a 3D-printed stand and they did not move during the experiment. 
In the human eye, the retina has alternating movement, caused predominantly by breathing and the heartbeat.
This motion can be a source of unintended damage to the delicate retinal cells if not accounted for.
For example, the retinal movement along the axial direction ($Z$-axis, perpendicular to the retina layer) caused by heartbeat and breathing is reported to be of the amplitude of 81$\pm$3.5 \textmu m with a frequency of approximately 1 Hz. 
Additionally, the amplitude of the retina's axial motion is recorded as 21.3$\pm$8 \textmu m while the subject is lying down to minimize the effects of head movements \cite{de2011heartbeat}. 

This work presents a novel approach for motion compensation in autonomous robotic eye surgery, explicitly targeting subretinal injections. 
Our method leverages real time acquisition and segmentation of small OCT volumes, called B$^{5}$-scans, to track retinal movement along the $Z$-axis. 
This allows us to dynamically adjust the position of the SHER robotic system's end-effector, maintaining the position of the surgical tools in relation to the retina (Fig. \ref{fig:breathing-phases}). 
To our knowledge, this is the first implementation of an OCT-based motion compensation technique in the context of autonomous robotic surgery. 
However, it is essential to note that this approach is an initial proof-of-concept. 
While our method successfully demonstrates the feasibility of real time motion compensation, it remains a first step towards more advanced and reliable techniques for automated subretinal interventions.
The contributions of this work consist of: 
\begin{itemize}
    \item The first implementation of an autonomous motion compensation system for robotic eye surgery, specifically targeting subretinal injections. This system uses real time OCT volume acquisition and segmentation, facilitated by small-volume scans termed B$^{5}$-scans. By tracking the motion of retinal layers in real time, our method enables precise control of the robotic end-effector to match retinal movements.
    \item We conducted validation experiments on \textit{ex vivo}, open-sky porcine eyes, simulating retinal motion through a sinusoidal pattern that mimics physiological movements such as respiration. Our evaluation focused on maintaining a consistent tool-to-retina distance and included performing subretinal injections.
\end{itemize}
The remainder of this paper is presented as follows. Section \ref{sec:methods} presents the proposed methods. Section \ref{sec:experiments} describes experimental setup and procedures. Results are provided in Section \ref{sec:results} and discussed in Section \ref{sec:discussion}. Section \ref{sec:conclusions} concludes the paper.

\begin{figure*}
    \centering
    \includegraphics[width=0.9\linewidth]{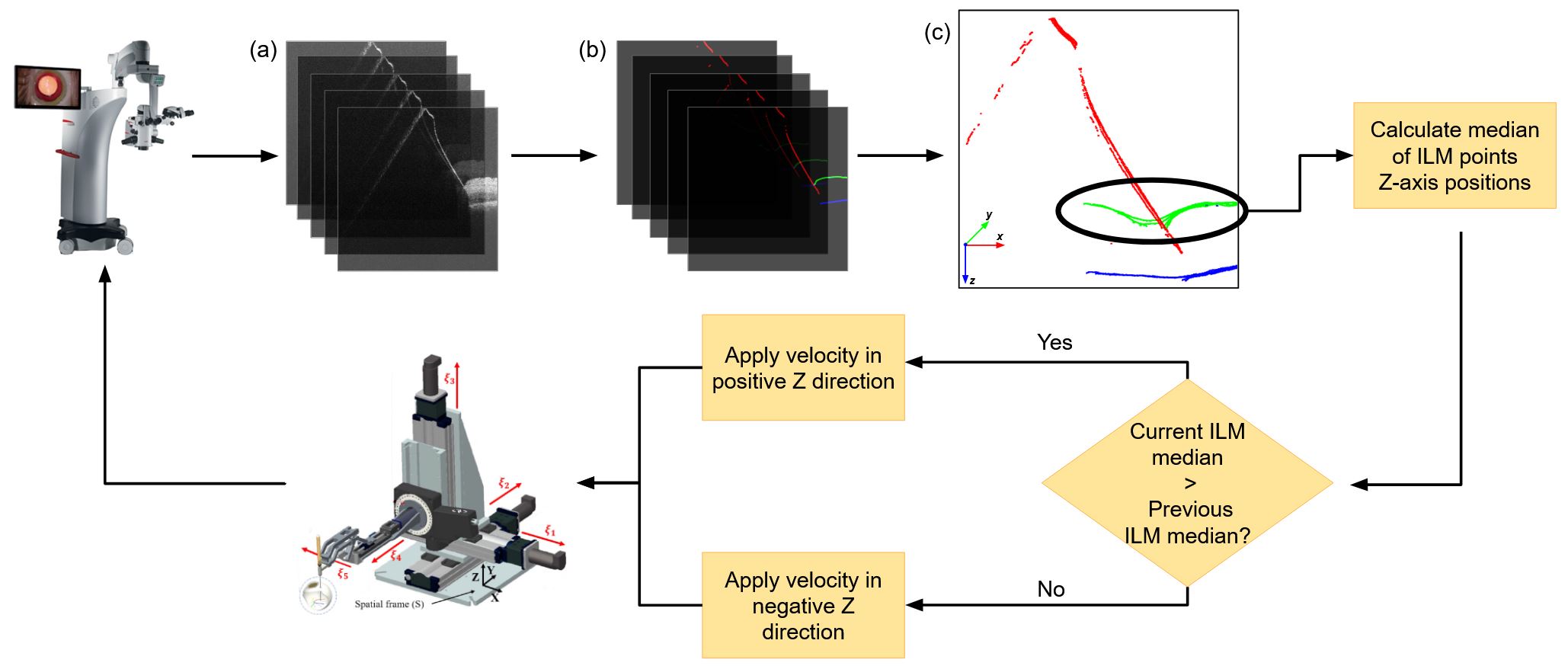}
    \caption{Our pipeline starts by acquiring B$^{5}$-scans (a), segmenting them using a deep neural network (b) and generating a surface point cloud based on the segmentation results (c). The median Z-axis position of the ILM points (green) in the point cloud is calculated and compared to the previous scan. Based on their difference, a velocity is applied to the robot end-effector and the process is repeated until it is terminated.}
    \label{fig:method-overview}
\end{figure*}

%%%%%%%%%%%%%%%%%%%%%%%%%%%%%%%%%%%%%%%%%%%%%%

\section{METHODS} \label{sec:methods}
Our motion compensation is based on the real time processing of OCT volumes. As depicted in Fig. \ref{fig:method-overview}, the incoming OCT volumes are segmented and point clouds of the ILM, RPE, and needle surfaces are generated from the segmentation results. 
The direction of the retina's motion is determined by comparing the median $Z$-axis position of the layer points to the previous point cloud.
A velocity is applied to the robot end-effector in the positive or negative $Z$ direction in accordance with the retinal motion, allowing the needle to be maintained at the same relative position to the retina.
\subsection{B$^{5}$-scans}
Since OCT microscope systems with real time C-scan acquisition capabilities are not yet commercially available, we use small volume scans, which we call B$^{5}$-scans, instead of large C-scans for imaging the surgical area. 
By combining five equally spaced B-scans of a 0.1~$\times$~4 mm scan area, we can create a B$^{5}$-scan.
Because of the small number of B-scans, B$^{5}$-scans are acquired in approximately 0.1 seconds. 

For comparison, using the Leica Proveo 8 with the EnFocus OCT imaging system acquiring a conventional large C-scan (4~$\times$~4 mm scan area, 200~$\times$~1000~$\times$~1024 pixel resolution) takes 6 seconds.
The fast acquisition times provide real time information necessary for robotic applications. 

\subsection{OCT Segmentation Network}
The positions of the retinal layers need to be known to calculate retinal motion. 
We used an OCT B-scan segmentation network previously introduced in \cite{arikan2024}, where we described our training data and procedure.
This network segments the top layer surface of the ILM, RPE, and the needle. 
This work presents the network parameters in Table I, not included in our previous work.

\begin{table}[h]
\centering
\caption{OCT B-scan segmentation network parameters}
\label{tab:model-param-table}
\begin{tabular}{lc}
\textbf{Parameter}     & \textbf{Value}              \\ \hline
\textbf{spatial\_dims} & 2                           \\
\textbf{in\_channels}  & 1                           \\
\textbf{out\_channels} & 4                           \\
\textbf{channels}      & (16, 32, 64, 128, 256, 512) \\
\textbf{strides}       & (2, 2, 2, 2, 2)             \\
\textbf{kernel\_size}  & 3                           \\ \hline
\end{tabular}
\end{table}

\subsection{Surface Point Cloud Generation}
As this work is an addition to the autonomous robotic subretinal injection pipeline in \cite{arikan2024}, we chose to use the generated surface point cloud for calculating the layer depth. 
This reduces the number of points significantly while potentially removing mislabeled pixels.

As we are only interested in the topmost surfaces of the needle, ILM, and RPE layers, we create a point cloud by taking the first occurrence of each class along the vertical axis. 
A more detailed explanation of this procedure can be found in \cite{arikan2024}.

\subsection{Retina Motion Compensation}
To calculate the motion of the retina in the $Z$ direction, we use the median $Z$-axis position of the ILM layer points from the point cloud generated in the previous step. 
The choice of the median, rather than the mean, is deliberate: it minimizes the impact of outliers that could lead to inaccurate conclusions. 
Focusing on the median $Z$-axis position ensures a more robust representation of the ILM's position, despite potential noise or extreme values in the data.

To estimate the direction of the retina’s motion, we compare this median $Z$-axis position to the value from the previous B$^{5}$-scan. 
Suppose the current median $Z$-axis position is greater than the previous one. In that case, the retina is moving downward along the $Z$-axis and the opposite is true if the $Z$-axis position is smaller than the last.
Comparing the positions of the incoming scans with the previous one, instead of the first scan, overcomes the limitations of intentional or unintentional microscope motion, for example by adjusting the scan area for better visualization. 

Based on the direction of the retinal motion, we apply a constant velocity in the same direction as the robot end-effector along the $Z$-axis. 
The applied velocity is updated approximately every 0.11 seconds after the acquisition and processing of B$^{5}$-scans.

\begin{figure*}
    \centering
    \includegraphics[ width=0.9\linewidth]{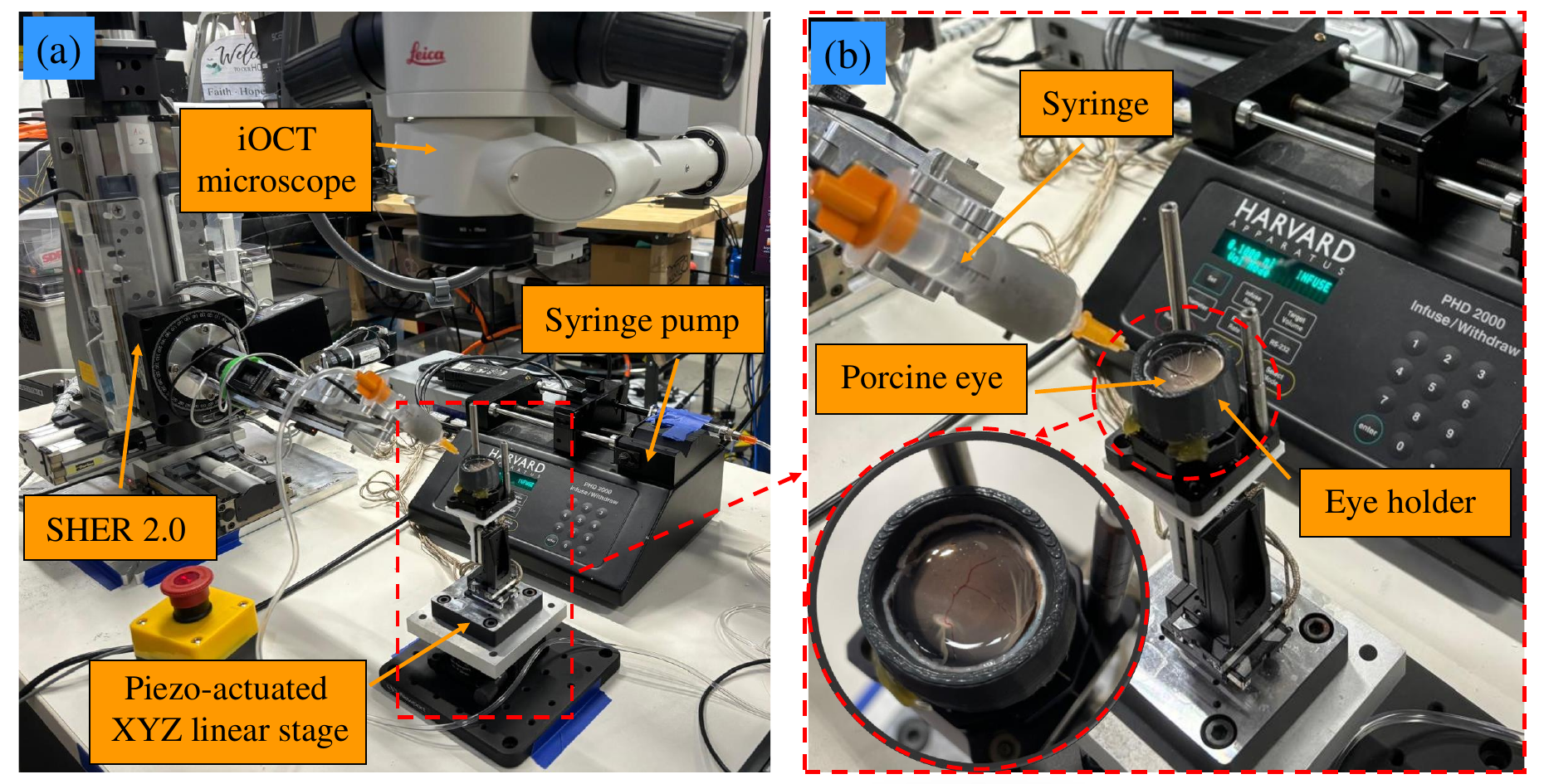}
    \caption{The experimental setup includes, (a) the SHER 2.0, an OCT microscope, a piezo-actuated linear stage to generate a sinusoidal motion along the $Z$ direction (up and down) simulating effects of human breathing on the eye, a syringe pump, a syringe with a 42 gauge needle, and (b) a closer view of the open sky porcine eye in the 3D printed holder.}
    \label{fig:System_setup_ISMR}
\end{figure*}

% \begin{figure}
%     \centering
%     \includegraphics[ width= 1.0\linewidth]{figures/pig_eye_preparaton_steps.pdf}
%     \caption{The preparation steps of the porcine eye for the experiment. The unprocessed ex vivo porcine eye (a) is cleaned by excision of the surrounding extraocular muscles (b), then bisected to remove the anterior segment including cornea, pupil, and lens, thereby exposing the retina and vitreous (c) and placed in the 3D printed holder attached to the piezo-actuated linear stage (d).}
%     \label{fig:eye-prep}
% \end{figure}

%%%%%%%%%%%%%%%%%%%%%%%%%%%%%%%%%%%%%%%%%%%%%%%%%%%%
\begin{figure}[h!]
    \centering
    \includegraphics[width=\linewidth]{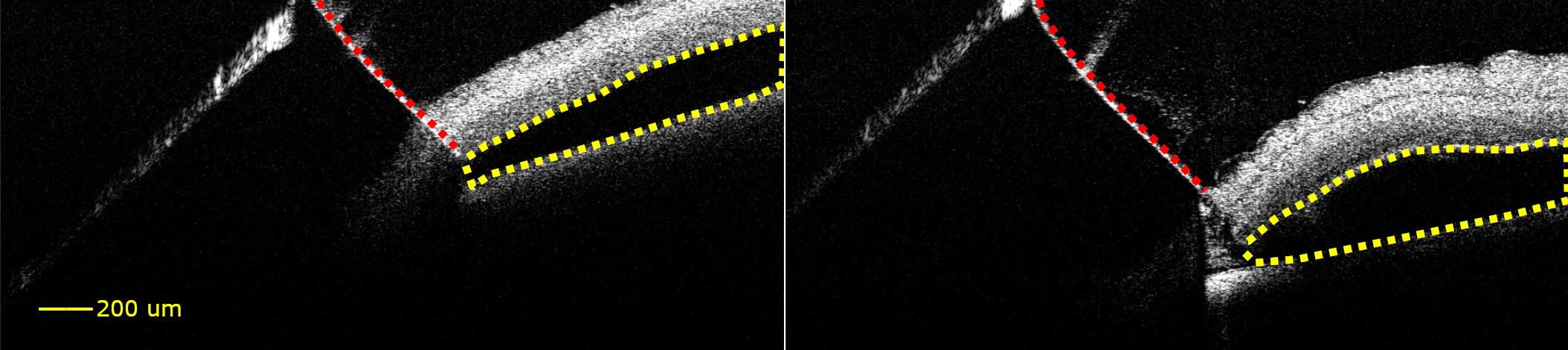}
    \caption{Examples of the retinal detachment caused by the bleb (yellow) creation after successful subretinal injection. The needle is shown in red.}
    \label{fig:succesful-bleb}
\end{figure}

\section{EXPERIMENTS} \label{sec:experiments}
\subsection{Experimental Setup}
Fig. \ref{fig:System_setup_ISMR} presents the experimental setup, which includes the SHER 2.0 whose joint-level velocities are controlled by a Galil motion controller (Galil 4088, Galil, Rocklin, CA, USA), an OCT surgical microscope with a 5 mm imaging depth in tissue (Proveo 8, Leica Microsystems, Germany), a syringe pump (PHD2000, Harvard Apparatus, USA), a piezo-actuated linear stage (Q-Motion Stages, PI Motion and Positioning, MA, USA), a syringe with a 42 gauge needle (INCYTO Co., South Korea), a 3D printed eye holder, and porcine eyes. The robot velocity controller is implemented in C++ using the CISST-SAW libraries \cite{deguet2008cisst}, and the deep learning algorithms are implemented in Python using PyTorch and MONAI \cite{Cardoso2022-ty} frameworks. All these components communicate via the Robot Operating System (ROS) through a TCP-IP connection.    

\subsection{Experimental Procedure}
In our experiments, we simulated the natural up-and-down motion of the retina that occurs due to respiration or heartbeats during surgeries using a sinusoidal function. 
By employing a linear stage, we moved the eye along the $Z$-axis in a controlled and predictable manner, closely following a perfect sine wave. 
This approach mirrors the regularity of natural breathing patterns and is like the predictable motion observed during ventilator-induced breathing under general anesthesia.
The sinusoidal function has a period of 5 seconds, corresponding to the average human breathing cycle \cite{Chourpiliadis2024}. 

For our initial experiments, we tested amplitudes of 25, 50, and 100 \textmu m to replicate varying degrees of retinal motion. 
In these preliminary tests, the needle was positioned at a random point above the retina, without contact, and we aimed to maintain the same distance between the needle and the retinal surface as in the starting position.

The velocity that was applied to the robot end-effector was set equal to the velocity of the retinal motion, which was calculated based on the amplitude and period of its sinusoidal displacement. 
Given that the sinusoidal motion has a period of 5 seconds and that the retina covers a total distance equivalent to four times the amplitude in each cycle, we calculate retinal velocity (V$_{retina}$) as follows:

\begin{equation}
V_{\text{retina}} = \frac{4 \times \text{Amplitude}}{\text{Period}}
\end{equation}

\begin{figure*}[h]
\begin{subfigure}[t]{0.5\textwidth}
    \includegraphics[width=\textwidth]{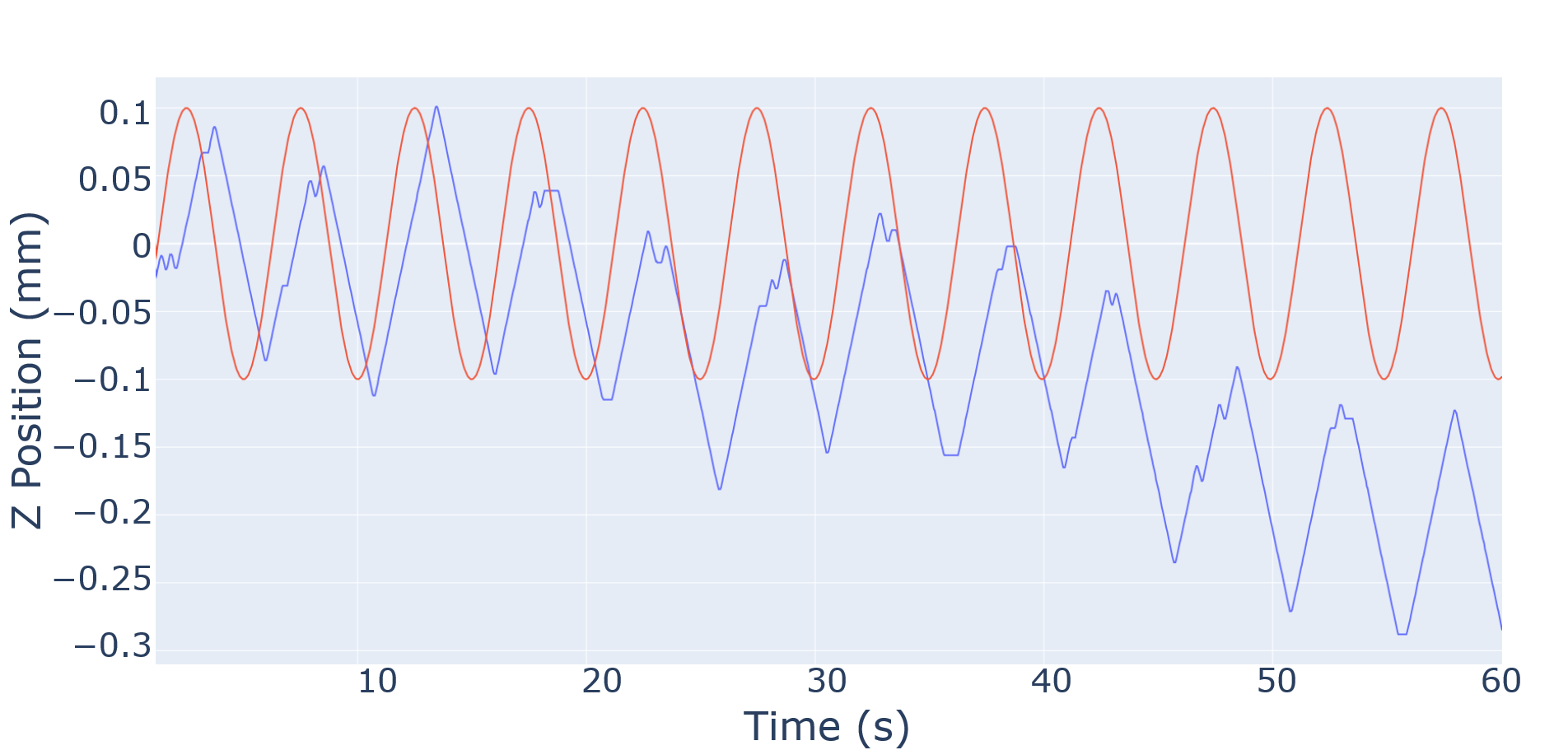}
    \caption{100 \textmu m amplitude retina motion, 0.8 mm/s robot velocity}
\end{subfigure}\hspace{\fill} % maximize horizontal separation
\begin{subfigure}[t]{0.5\textwidth}
    \includegraphics[width=\linewidth]{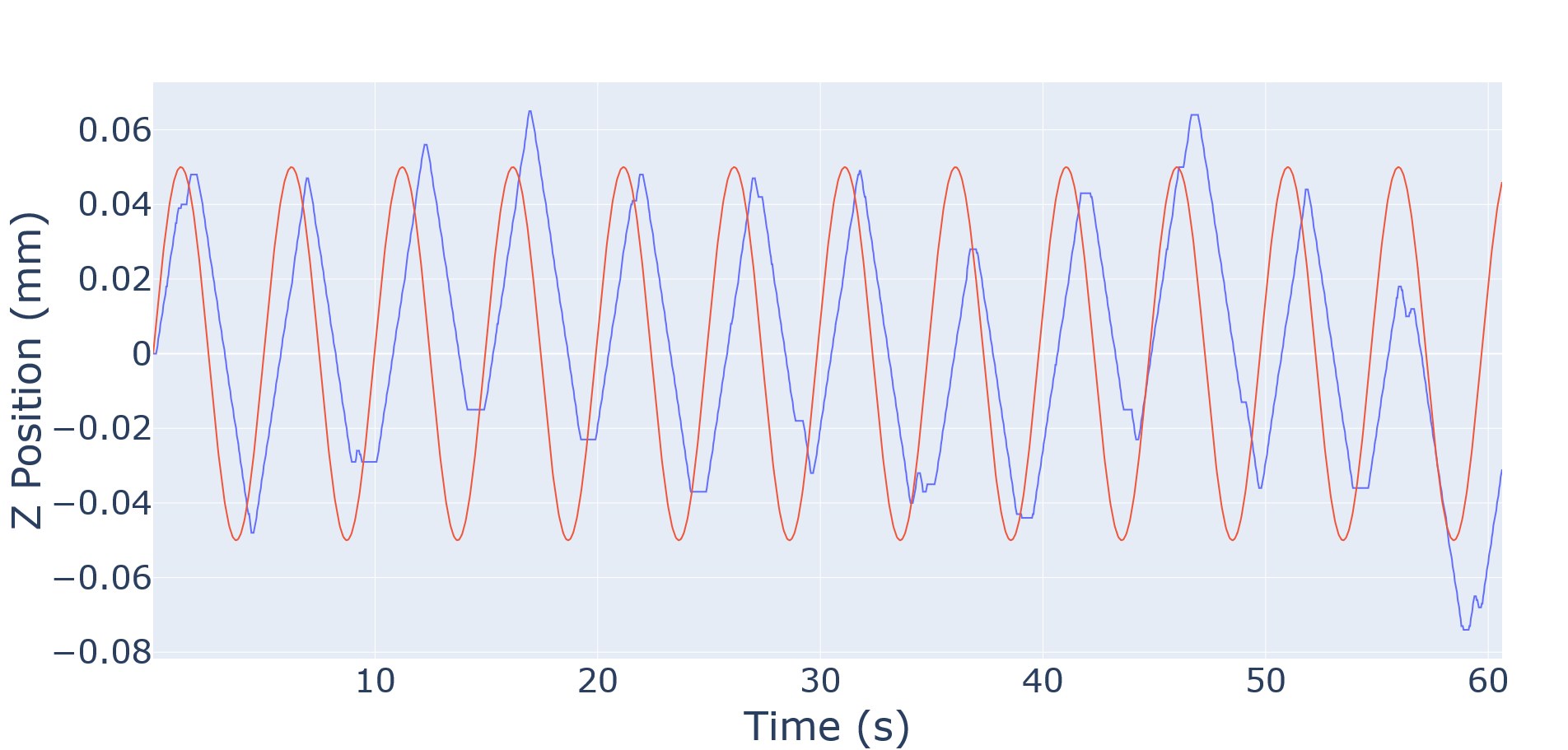}
    \caption{50 \textmu m amplitude retina motion, 0.4 mm/s robot velocity}
\end{subfigure}
\begin{subfigure}[t]{0.5\textwidth}
    \includegraphics[width=\linewidth]{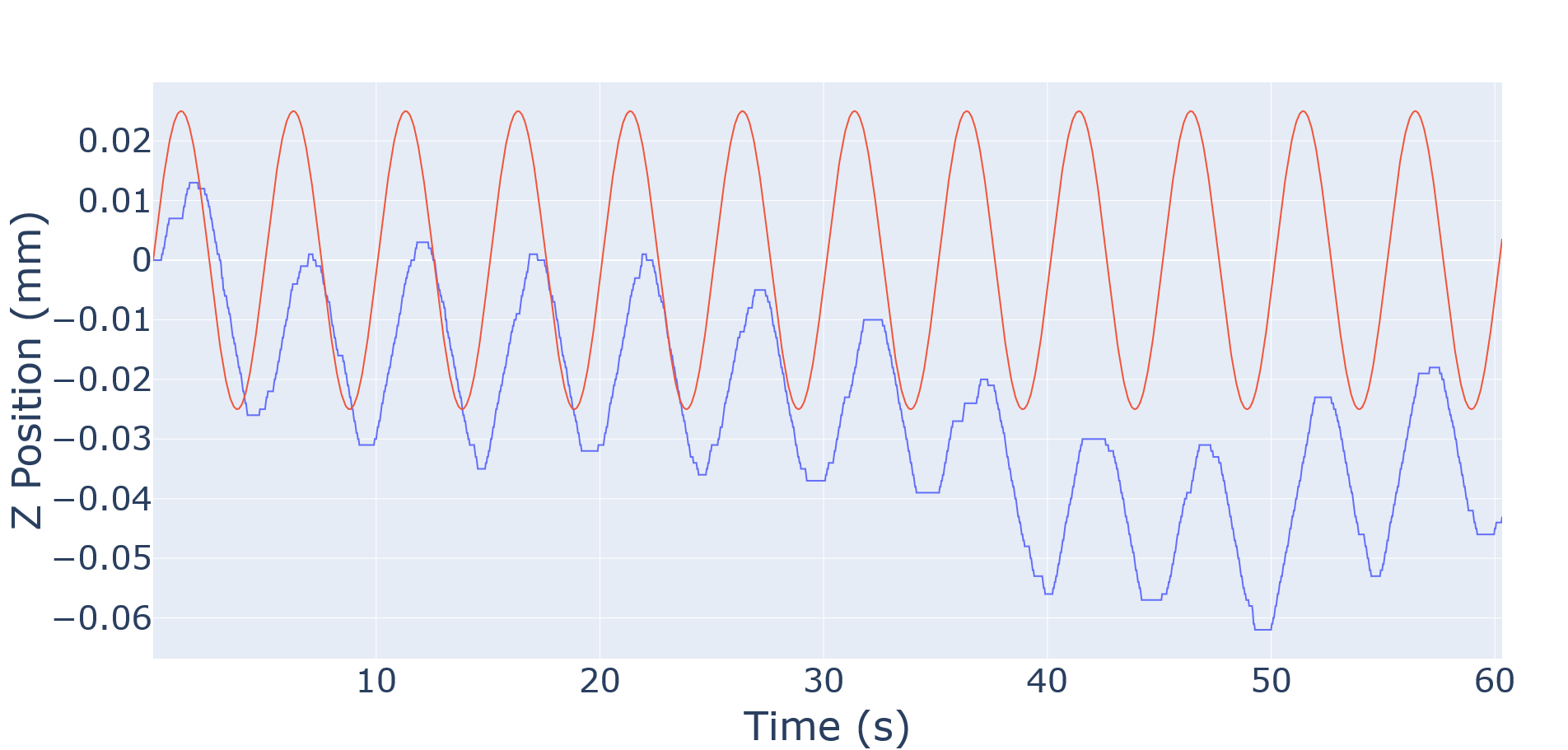}
    \caption{25 \textmu m amplitude retina motion, 0.2 mm/s robot velocity}
\end{subfigure}\hspace{\fill} % maximize horizontal separation
\begin{subfigure}[t]{0.5\textwidth}
    \includegraphics[width=\linewidth]{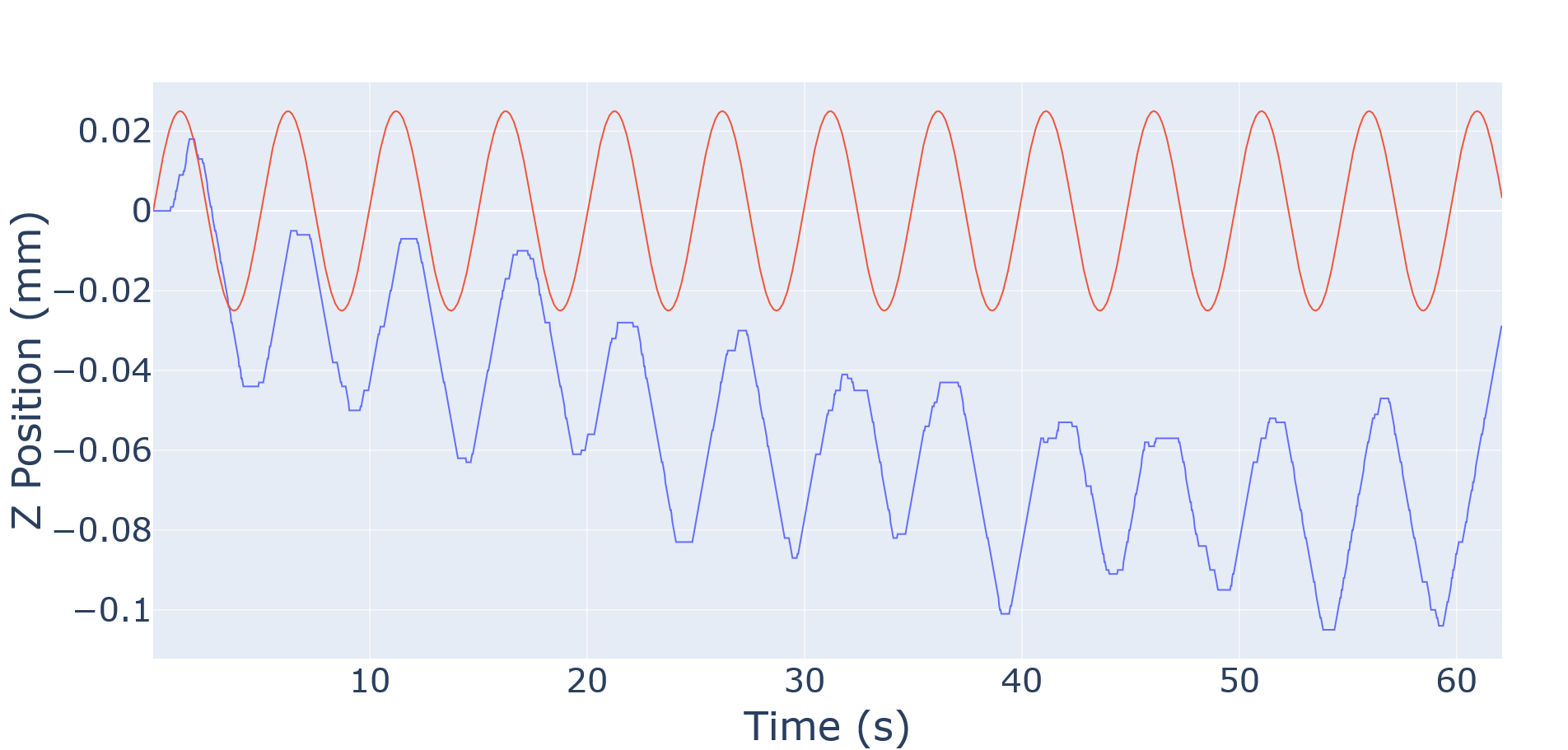}
    \caption{25 \textmu m amplitude retina motion, 0.3 mm/s robot velocity}
\end{subfigure}
\caption{Z-axis positions of the needle tip (blue) and linear stage controller (orange) over time.}
\label{fig:constant-dist-results}
\end{figure*}

Based on this formula the velocity of the 25, 50, 100 \textmu m amplitude motions are 0.2, 0.4 and 0.8 mm/s respectively. 
Although the 0.2 mm/s velocity is theoretically correct for a 25 \textmu m amplitude motion, we also tested applying a slightly higher velocity of 0.3 mm/s to the robot end-effector. 

In the second experiment, we set the eye motion amplitude to 100 \textmu m and positioned the needle above the retina. 
Using the method outlined in our previous paper \cite{arikan2024}, we performed needle insertions and activated the breathing compensation described here once the target insertion depth was achieved. 
We moved the robot end-effector with a 0.8 mm/s velocity based on the results of our previous experiment.
Following insertion, a 0.1 ml fluid injection was administered using a syringe pump with an infusion rate of 1 ml/min to evaluate the insertion’s quality. At the same time, both the eye and needle were in motion. 
This procedure was repeated for seven separate injections.
We considered the injection successful if the injected water flowed under the retina causing retinal detachment and retinal bleb formation. Examples of successful injections can be seen in Fig. \ref{fig:succesful-bleb}.

\section{RESULTS} \label{sec:results}
\begin{figure}
    \centering
    \includegraphics[width=\linewidth]{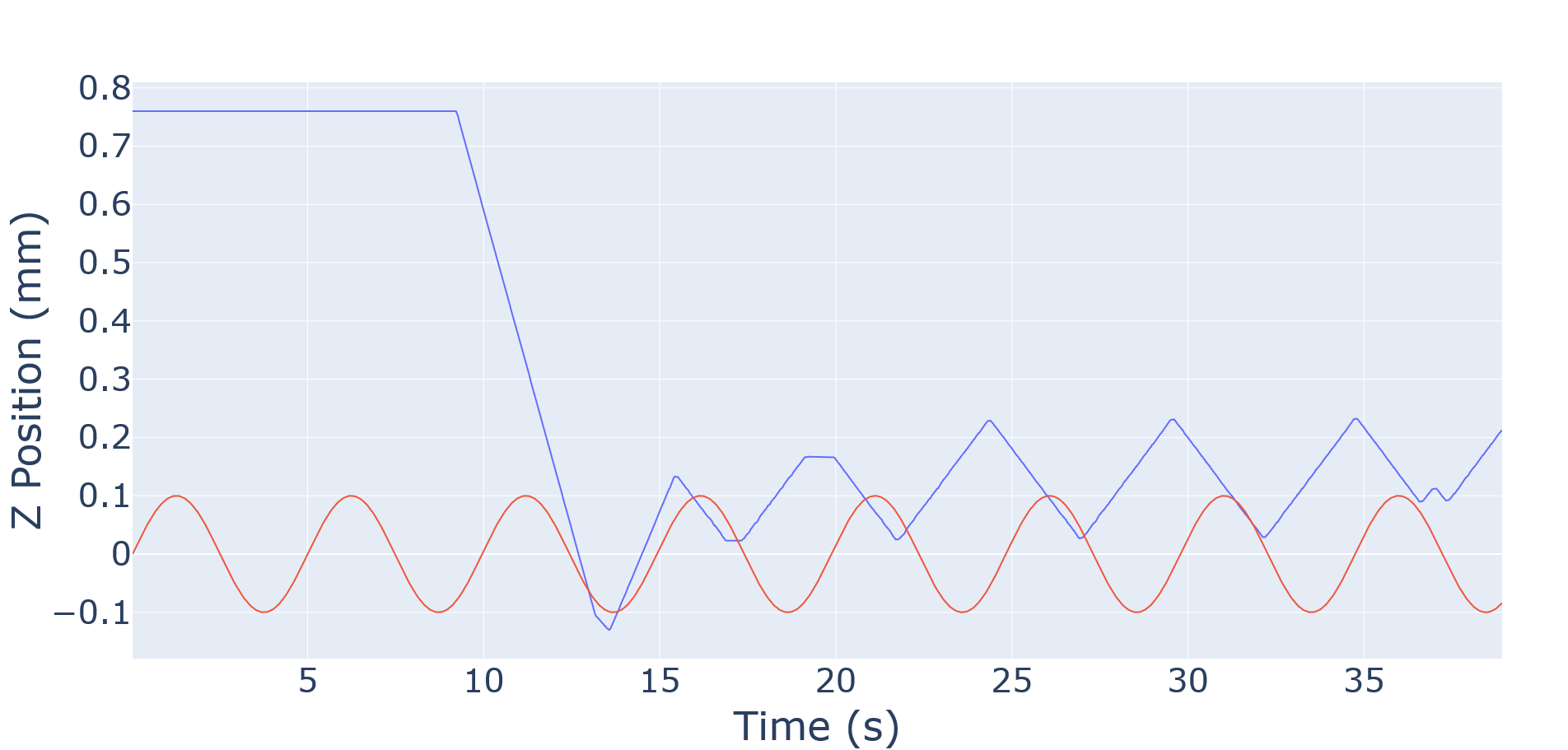}
    \caption{Z-axis position of the needle tip (blue) and linear stage controller (orange) during a subretinal injection procedure.}
    \label{fig:breathing-insertion}
\end{figure}

In Fig. \ref{fig:constant-dist-results}, we present the $Z$-axis positions of the linear stage controller and needle over time during our first experiment, where we tried to maintain a constant distance between the needle tip and the retina for 60 seconds. 
Both values are zeroed at their starting positions to allow direct visual comparison.
In Fig. \ref{fig:constant-dist-results} (c) and (d), we show results from the 25 \textmu m amplitude motion experiments with 0.2 and 0.3 mm/s robot velocities, respectively. 

For the 100 \textmu m amplitude motion, we see approximately a 200 \textmu m vertical deviation that starts around the 40-second mark. The 50 \textmu m amplitude motion performs better at following the linear stages motion. 

In Fig. \ref{fig:constant-dist-results} (c) we can see that the theoretically correct 0.2 mm/s end-effector velocity does not have a similar amplitude to the actual linear stage motion. 
Although using 0.3 mm/s end-effector velocity, we can better match the amplitude of the reference motion. This experiment also has ahigh vertical shift, with a maximum of 80 \textmu m. 

Figure \ref{fig:breathing-insertion} presents a graph illustrating the needle positions and the linear stage controller over time, depicting a typical result from our second experiment. 
In this experiment, we performed autonomous subretinal injections while the linear stage moved in a controlled sinusoidal pattern to simulate physiological eye movement. 
The graph shows how the robotic system adapted to maintain needle positioning during this dynamic procedure.
Out of the seven injections that we performed, four of them were successful in creating a bleb based on our criteria.

Notably, the linear stage motion (orange line) is the sinusoidal motion that the linear stage is programmed to move. 
The values seen there do not directly correspond to the real-life motion of the eye, especially when the needle is inserted into the retina. 
We discuss this further in the following sections of this paper.

%%%%%%%%%%%%%%%%%%%%%%%%%%%%%%%%%%%%%%%%%%%%%%%%%%%%%\

\section{DISCUSSION AND FUTURE WORK} \label{sec:discussion}
Our results demonstrate that our robotic system can match the sinusoidal motion of the eye on the Z-axis. 
For the 100 and 50 \textmu m amplitude motions, our robot can match their movement. 
But for the 25 \textmu m amplitude motion, we see that the 0.2 mm/s end-effector velocity does not match the motion of the linear stage. 
This could be due to limitations of our robot's ability to move at low velocities. This problem is solved by increasing the applied velocity to 0.3 mm/s.

One of the main limitations observed is a phase shift between the needle’s motion and the retinal tissue. This latency arises from our position correction method, which compares each incoming B${^5}$-scan with the previously processed scan. 
Consequently, the system continuously adjusts the robot’s position based on the prior state of the retina rather than its real time position. 
As a result, each corrective movement is slightly delayed, leading to the phase lag seen in the graphs in Fig. \ref{fig:constant-dist-results} and the OCT images in Fig. \ref{fig:breathing-phases}. 
This lag affects the precision of needle tracking. It suggests that our system may struggle to maintain synchronization with rapidly moving retinal tissue.

A secondary issue noted in the results is the gradual vertical drift in the needle’s position relative to the retina. 
Over time, the needle tip deviates from its starting position, shifting either upwards or downwards.
This phenomenon is especially prevalent in the 25 \textmu m amplitude retina motion examples (Fig. \ref{fig:constant-dist-results} (c) and (d)). 
% During our secondary experiments with subretinal injections the vertical shift also caused the needle to move away from the target depth, leading to water being injected into the vitreous. 
Additionally, there are examples in our results where the needle velocity is zero (position line is horizontal) or changes direction multiple times in a short time.
This phenomenon and the vertical drift are caused by errors that happen during the acquisition and processing of scans. 
Although the average acquisition time of the OCT scans is 0.1 seconds, there are cases where the acquisition can take longer, causing a delayed update of the velocity, or shorter, causing multiple updates in a short time. 
Furthermore, wrongly segmented scans can lead to instances where the robot can receive a couple of velocity updates in the opposite direction followed by continued correct velocity updates, causing noisy or horizontal position lines.
Accumulating the incorrect velocity changes leads to the drift seen in the results.

To address the latency and vertical drift issues, we propose incorporating a predictive motion model that can anticipate the quasi-periodic motion of the retina caused by the patient’s breathing and heartbeat. 
For instance, by integrating a Kalman filter the system can predict the retina’s upcoming position based on previous motion patterns. This allows it to adjust the needle’s movement proactively rather than reactively. 
A predictive model could reduce the phase lag by anticipating shifts and potentially eliminate vertical drift by consistently recalibrating to account for any inaccuracies in position tracking.

\begin{figure}
    \centering
    \includegraphics[width=0.8\linewidth]{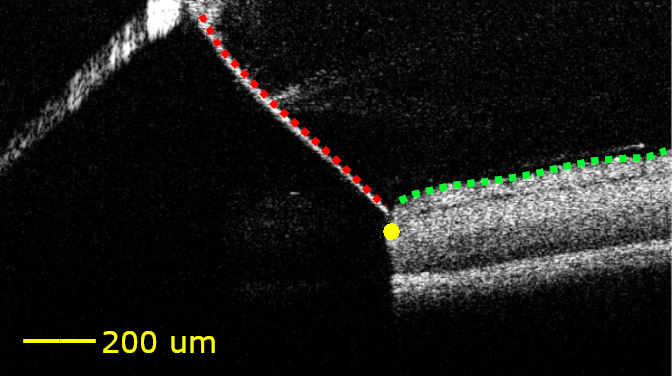}
    \caption{OCT B-scan image captured at the 20-second mark of the insertion shown in Figure \ref{fig:breathing-insertion}. A yellow circle marks the needle tip, the needle is highlighted in red, and the ILM layer of the retina is highlighted in green.}
    \label{fig:insertion-oct}
\end{figure}

Another observation is the altered retina motion associated with the needle insertion. 
Looking at Fig. \ref{fig:breathing-insertion}, the needle leaves the tissue by around 100 \textmu m when the needle and linear stage motions are out of sync.
However, when looking at the OCT scans, we realize that the needle tip remains in the retina and the horizontal shift is much less significant.  
When the needle penetrates the retina, it likely induces surface adhesion and resistance within the tissue, restricting free movement. 
This effect is particularly evident during downward movements, where the needle’s presence within the tissue exerts a tethering force that impedes motion. 
In Fig. \ref{fig:insertion-oct}, we show an OCT B-scan at the 20-second mark where the needle tip is still in contact with the retina.
According to the graph, the needle should be outside of the retinal tissue at this point.
Therefore, when tool-tissue interactions are present, the linear stage controller’s position does not fully represent the position or motion.

%%%%%%%%%%%%%%%%%%%%%%%%%%%%%%%%%%%%%%%%%%%%%%%%%%%%%%%%%
\section{CONCLUSION} \label{sec:conclusions}

In this work, we presented our initial method and experiments for extending our previously developed autonomous robotic subretinal injection pipeline with motion compensation capabilities along the $Z$-axis. 
Based on real time acquisition and segmentation of small 3D volumes, called B$^{5}$-scans, we calculate the depth of the ILM layer and compare it with the previously computed values. 
We then move the robot end-effector to keep the needle tip in the same position relative to the retina. 

Our method shows the potential of our robotic system to compensate for simulated motion like motion caused by respiration or circulation. However, some aspects should be further examined, such as the interactions of moving tissue and surgical tools and the implementation of predictive methods for estimating retinal motion for more accurate compensation.
These are potential future work areas. 
%%%%%%%%%%%%%%%%%%%%%%%%%%%%%%%%%%%%%%%%%%%%%%%%%%%%%%%%%%%%%%%%%%%%%%%%%%%%%%%%

% \section*{ACKNOWLEDGMENT}
% Research to Prevent Blindness, New York, New York, USA, and gifts by the J. Willard and Alice S. Marriott Foundation, the Gale Trust, Mr. Herb Ehlers, Mr. Bill Wilbur, Mr. and Mrs. Rajandre Shaw, Ms. Helen Nassif, Ms Mary Ellen Keck, Don and Maggie Feiner, Dick and Gretchen Nielsen.

%%%%%%%%%%%%%%%%%%%%%%%%%%%%%%%%%%%%%%%%%%%%%%%%%%%%%%%%%%%%%%%%%%%%%%%%%%%%%%%%
\newpage
\bibliographystyle{IEEEtran}
\bibliography{root}

\begin{thebibliography}{10}
\providecommand{\url}[1]{#1}
\csname url@samestyle\endcsname
\providecommand{\newblock}{\relax}
\providecommand{\bibinfo}[2]{#2}
\providecommand{\BIBentrySTDinterwordspacing}{\spaceskip=0pt\relax}
\providecommand{\BIBentryALTinterwordstretchfactor}{4}
\providecommand{\BIBentryALTinterwordspacing}{\spaceskip=\fontdimen2\font plus
\BIBentryALTinterwordstretchfactor\fontdimen3\font minus \fontdimen4\font\relax}
\providecommand{\BIBforeignlanguage}[2]{{%
\expandafter\ifx\csname l@#1\endcsname\relax
\typeout{** WARNING: IEEEtran.bst: No hyphenation pattern has been}%
\typeout{** loaded for the language `#1'. Using the pattern for}%
\typeout{** the default language instead.}%
\else
\language=\csname l@#1\endcsname
\fi
#2}}
\providecommand{\BIBdecl}{\relax}
\BIBdecl

\bibitem{wong2008natural}
T.~Wong, U.~Chakravarthy, R.~Klein, P.~Mitchell, G.~Zlateva, R.~Buggage, K.~Fahrbach, C.~Probst, and I.~Sledge, ``The natural history and prognosis of neovascular age-related macular degeneration: a systematic review of the literature and meta-analysis,'' \emph{Ophthalmology}, vol. 115, no.~1, pp. 116--126, 2008.

\bibitem{kawasaki2010prevalence}
R.~Kawasaki, M.~Yasuda, S.~J. Song, S.-J. Chen, J.~B. Jonas, J.~J. Wang, P.~Mitchell, and T.~Y. Wong, ``The prevalence of age-related macular degeneration in asians: a systematic review and meta-analysis,'' \emph{Ophthalmology}, vol. 117, no.~5, pp. 921--927, 2010.

\bibitem{mitchell1995prevalence}
P.~Mitchell, W.~Smith, K.~Attebo, and J.~J. Wang, ``Prevalence of age-related maculopathy in australia: the blue mountains eye study,'' \emph{Ophthalmology}, vol. 102, no.~10, pp. 1450--1460, 1995.

\bibitem{klein1999prevalence}
R.~Klein, B.~E. Klein, and K.~J. Cruickshanks, ``The prevalence of age-related maculopathy by geographic region and ethnicity,'' \emph{Progress in retinal and eye research}, vol.~18, no.~3, pp. 371--389, 1999.

\bibitem{klaver2001incidence}
C.~C. Klaver, J.~J. Assink, R.~van Leeuwen, R.~C. Wolfs, J.~R. Vingerling, T.~Stijnen, A.~Hofman, and P.~T. de~Jong, ``Incidence and progression rates of age-related maculopathy: the rotterdam study,'' \emph{Investigative ophthalmology \& visual science}, vol.~42, no.~10, pp. 2237--2241, 2001.

\bibitem{wong2014global}
W.~L. Wong, X.~Su, X.~Li, C.~M.~G. Cheung, R.~Klein, C.-Y. Cheng, and T.~Y. Wong, ``Global prevalence of age-related macular degeneration and disease burden projection for 2020 and 2040: a systematic review and meta-analysis,'' \emph{The Lancet Global Health}, vol.~2, no.~2, pp. e106--e116, 2014.

\bibitem{finger2020anti}
R.~P. Finger, V.~Daien, B.~M. Eldem, J.~S. Talks, J.-F. Korobelnik, P.~Mitchell, T.~Sakamoto, T.~Y. Wong, K.~Pantiri, and J.~Carrasco, ``Anti-vascular endothelial growth factor in neovascular age-related macular degeneration--a systematic review of the impact of anti-vegf on patient outcomes and healthcare systems,'' \emph{BMC ophthalmology}, vol.~20, pp. 1--14, 2020.

\bibitem{day2011ocular}
S.~Day, K.~Acquah, P.~Mruthyunjaya, D.~S. Grossman, P.~P. Lee, and F.~A. Sloan, ``Ocular complications after anti--vascular endothelial growth factor therapy in medicare patients with age-related macular degeneration,'' \emph{American journal of ophthalmology}, vol. 152, no.~2, pp. 266--272, 2011.

\bibitem{rakoczy2017gene}
E.~P. Rakoczy, ``Gene therapy for the long term treatment of wet amd,'' \emph{The Lancet}, vol. 390, no. 10089, pp. 6--7, 2017.

\bibitem{zhao2017development}
C.~Zhao, N.~C. Boles, J.~D. Miller, S.~Kawola, S.~Temple, R.~J. Davis, and J.~H. Stern, ``Development of a refined protocol for trans-scleral subretinal transplantation of human retinal pigment epithelial cells into rat eyes,'' \emph{Journal of Visualized Experiments: JoVE}, no. 126, 2017.

\bibitem{ochakovski2017retinal}
G.~A. Ochakovski, K.~U. Bartz-Schmidt, and M.~D. Fischer, ``Retinal gene therapy: surgical vector delivery in the translation to clinical trials,'' \emph{Frontiers in neuroscience}, vol.~11, p. 174, 2017.

\bibitem{irigoyen2022subretinal}
C.~Irigoyen, A.~Amenabar~Alonso, J.~Sanchez-Molina, M.~Rodr{\'\i}guez-Hidalgo, A.~Lara-L{\'o}pez, and J.~Ruiz-Ederra, ``Subretinal injection techniques for retinal disease: a review,'' \emph{Journal of clinical medicine}, vol.~11, no.~16, p. 4717, 2022.

\bibitem{riviere2000study}
C.~N. Riviere and P.~S. Jensen, ``A study of instrument motion in retinal microsurgery,'' in \emph{Proceedings of the 22nd Annual International Conference of the IEEE Engineering in Medicine and Biology Society (Cat. No. 00CH37143)}, vol.~1.\hskip 1em plus 0.5em minus 0.4em\relax IEEE, 2000, pp. 59--60.

\bibitem{uneri2010new}
A.~{\"U}neri, M.~A. Balicki, J.~Handa, P.~Gehlbach, R.~H. Taylor, and I.~Iordachita, ``New steady-hand eye robot with micro-force sensing for vitreoretinal surgery,'' in \emph{2010 3rd IEEE RAS \& EMBS International Conference on Biomedical Robotics and Biomechatronics}.\hskip 1em plus 0.5em minus 0.4em\relax IEEE, 2010, pp. 814--819.

\bibitem{rahimy2013robot}
E.~Rahimy, J.~Wilson, T.~Tsao, S.~Schwartz, and J.~Hubschman, ``Robot-assisted intraocular surgery: development of the iriss and feasibility studies in an animal model,'' \emph{Eye}, vol.~27, no.~8, pp. 972--978, 2013.

\bibitem{van2009design}
L.~van~den Bedem, R.~Hendrix, N.~Rosielle, M.~Steinbuch, and H.~Nijmeijer, ``Design of a minimally invasive surgical teleoperated master-slave system with haptic feedback,'' in \emph{2009 International Conference on Mechatronics and Automation}.\hskip 1em plus 0.5em minus 0.4em\relax IEEE, 2009, pp. 60--65.

\bibitem{gijbels2014experimental}
A.~Gijbels, E.~B. Vander~Poorten, B.~Gorissen, A.~Devreker, P.~Stalmans, and D.~Reynaerts, ``Experimental validation of a robotic comanipulation and telemanipulation system for retinal surgery,'' in \emph{5th IEEE RAS/EMBS International Conference on Biomedical Robotics and Biomechatronics}.\hskip 1em plus 0.5em minus 0.4em\relax IEEE, 2014, pp. 144--150.

\bibitem{jingjing2014design}
X.~Jingjing, H.~Long, S.~Lijun, and Y.~Yang, ``Design and research of a robotic aided system for retinal vascular bypass surgery,'' \emph{Journal of Medical Devices}, vol.~8, no.~4, p. 044501, 2014.

\bibitem{nasseri2013introduction}
M.~A. Nasseri, M.~Eder, S.~Nair, E.~Dean, M.~Maier, D.~Zapp, C.~P. Lohmann, and A.~Knoll, ``The introduction of a new robot for assistance in ophthalmic surgery,'' in \emph{2013 35th Annual International Conference of the IEEE Engineering in Medicine and Biology Society (EMBC)}.\hskip 1em plus 0.5em minus 0.4em\relax IEEE, 2013, pp. 5682--5685.

\bibitem{zhao2023human}
B.~Zhao, M.~Esfandiari, D.~E. Usevitch, P.~Gehlbach, and I.~Iordachita, ``Human-robot interaction in retinal surgery: A comparative study of serial and parallel cooperative robots,'' in \emph{2023 32nd IEEE International Conference on Robot and Human Interactive Communication (RO-MAN)}.\hskip 1em plus 0.5em minus 0.4em\relax IEEE, 2023, pp. 2359--2365.

\bibitem{esfandiari2024cooperative}
M.~Esfandiari, J.~W. Kim, B.~Zhao, G.~Amirkhani, M.~Hadi, P.~Gehlbach, R.~H. Taylor, and I.~Iordachita, ``Cooperative vs. teleoperation control of the steady hand eye robot with adaptive sclera force control: A comparative study,'' in \emph{2024 IEEE International Conference on Robotics and Automation (ICRA)}.\hskip 1em plus 0.5em minus 0.4em\relax IEEE, 2024, pp. 8209--8215.

\bibitem{Henry2025-dg}
R.~Henry, M.~Huber, A.~Mablekos-Alexiou, C.~Seneci, M.~Abdelaziz, H.~Natalius, L.~da~Cruz, and C.~Bergeles, ``Evaluating robotic approach techniques for the insertion of a straight instrument into a vitreoretinal surgery trocar,'' 2025.

\bibitem{zhou2019towards}
M.~Zhou, Q.~Yu, K.~Huang, S.~Mahov, A.~Eslami, M.~Maier, C.~P. Lohmann, N.~Navab, D.~Zapp, A.~Knoll \emph{et~al.}, ``Towards robotic-assisted subretinal injection: A hybrid parallel--serial robot system design and preliminary evaluation,'' \emph{IEEE Transactions on Industrial Electronics}, vol.~67, no.~8, pp. 6617--6628, 2019.

\bibitem{sommersperger2021real}
M.~Sommersperger, J.~Weiss, M.~Ali~Nasseri, P.~Gehlbach, I.~Iordachita, and N.~Navab, ``Real-time tool to layer distance estimation for robotic subretinal injection using intraoperative 4d oct,'' \emph{Biomedical Optics Express}, vol.~12, no.~2, pp. 1085--1104, 2021.

\bibitem{mach2022oct}
K.~Mach, S.~Wei, J.~W. Kim, A.~Martin-Gomez, P.~Zhang, J.~U. Kang, M.~A. Nasseri, P.~Gehlbach, N.~Navab, and I.~Iordachita, ``Oct-guided robotic subretinal needle injections: A deep learning-based registration approach,'' in \emph{2022 IEEE International Conference on Bioinformatics and Biomedicine (BIBM)}.\hskip 1em plus 0.5em minus 0.4em\relax IEEE, 2022, pp. 781--786.

\bibitem{dehghani2023robotic}
S.~Dehghani, M.~Sommersperger, P.~Zhang, A.~Martin-Gomez, B.~Busam, P.~Gehlbach, N.~Navab, M.~A. Nasseri, and I.~Iordachita, ``Robotic navigation autonomy for subretinal injection via intelligent real-time virtual ioct volume slicing,'' in \emph{2023 IEEE International Conference on Robotics and Automation (ICRA)}.\hskip 1em plus 0.5em minus 0.4em\relax IEEE, 2023, pp. 4724--4731.

\bibitem{arikan2024}
\BIBentryALTinterwordspacing
D.~Arikan, P.~Zhang, M.~Sommersperger, S.~Dehghani, M.~Esfandiari, R.~H. Taylor, M.~A. Nasseri, P.~Gehlbach, N.~Navab, and I.~Iordachita, ``Real-time deformation-aware control for autonomous robotic subretinal injection under ioct guidance,'' 2024. [Online]. Available: \url{https://arxiv.org/abs/2411.06557}
\BIBentrySTDinterwordspacing

\bibitem{de2011heartbeat}
R.~de~Kinkelder, J.~Kalkman, D.~J. Faber, O.~Schraa, P.~H. Kok, F.~D. Verbraak, and T.~G. van Leeuwen, ``Heartbeat-induced axial motion artifacts in optical coherence tomography measurements of the retina,'' \emph{Investigative ophthalmology \& visual science}, vol.~52, no.~6, pp. 3908--3913, 2011.

\bibitem{deguet2008cisst}
A.~Deguet, R.~Kumar, R.~Taylor, and P.~Kazanzides, ``The cisst libraries for computer assisted intervention systems,'' in \emph{MICCAI Workshop on Systems and Arch. for Computer Assisted Interventions, Midas Journal}, vol.~71, 2008.

\bibitem{Cardoso2022-ty}
M.~J. Cardoso, W.~Li, R.~Brown, N.~Ma, E.~Kerfoot, Y.~Wang, B.~Murrey, A.~Myronenko, C.~Zhao, D.~Yang, V.~Nath, Y.~He, Z.~Xu, A.~Hatamizadeh, A.~Myronenko, W.~Zhu, Y.~Liu, M.~Zheng, Y.~Tang, I.~Yang, M.~Zephyr, B.~Hashemian, S.~Alle, M.~Z. Darestani, C.~Budd, M.~Modat, T.~Vercauteren, G.~Wang, Y.~Li, Y.~Hu, Y.~Fu, B.~Gorman, H.~Johnson, B.~Genereaux, B.~S. Erdal, V.~Gupta, A.~Diaz-Pinto, A.~Dourson, L.~Maier-Hein, P.~F. Jaeger, M.~Baumgartner, J.~Kalpathy-Cramer, M.~Flores, J.~Kirby, L.~A.~D. Cooper, H.~R. Roth, D.~Xu, D.~Bericat, R.~Floca, S.~K. Zhou, H.~Shuaib, K.~Farahani, K.~H. Maier-Hein, S.~Aylward, P.~Dogra, S.~Ourselin, and A.~Feng, ``{MONAI}: An open-source framework for deep learning in healthcare,'' 2022.

\bibitem{Chourpiliadis2024}
C.~Chourpiliadis and A.~Bhardwaj, ``\BIBforeignlanguage{en}{Physiology, respiratory rate},'' in \emph{\BIBforeignlanguage{en}{{StatPearls}}}.\hskip 1em plus 0.5em minus 0.4em\relax Treasure Island (FL): StatPearls Publishing, Jan. 2024.

\end{thebibliography}
\end{document}